\documentclass[10pt,twocolumn,letterpaper]{article}

\usepackage[pagenumbers]{cvpr} 

\definecolor{cvprblue}{rgb}{0.21,0.49,0.74}
\usepackage[pagebackref,breaklinks,colorlinks,allcolors=cvprblue]{hyperref}

\def\paperID{20} 
\def\confName{Eval-FoMo}
\def\confYear{2025}

\title{A Vision Centric Remote Sensing Benchmark}
\makeatletter
 \newcommand{\printfnsymbol}[1]{%
  \textsuperscript{\@fnsymbol{#1}}%
 }
 \makeatother

\newcommand\blfootnote[1]{%
  \begingroup
  \renewcommand\thefootnote{}\footnote{#1}%
  \addtocounter{footnote}{-1}%
  \endgroup
}

\author{Abduljaleel Adejumo\thanks{equally contributing}\\
AMMI/AIMS\\
Senegal\\
{\tt\small aadejumo@aimsammi.org}
\and
Faegheh Yeganli\printfnsymbol{1}\\
University of British Columbia\\
Canada\\
{\tt\small fyeganli@mail.ubc.ca}
\and
Clifford Broni-bediako\\
RIKEN AIP\\
Japan\\
{\tt\small clifford.broni-bediako@riken.jp}
\and
Aoran Xiao\\
RIKEN AIP\\
Japan\\
{\tt\small aoran.xiao@riken.jp}
\and
Naoto Yokoya\thanks{equally advising}\\
RIKEN AIP, the University of Tokyo\\
Japan\\
{\tt\small yokoya@k.u-tokyo.ac.jp}
\and
Mennatullah Siam\printfnsymbol{2}\\
University of British Columbia\\
Canada\\
{\tt\small mennatullah.siam@ubc.ca}
}
\begin{document}

\maketitle

\begin{abstract}
Multimodal Large Language Models (MLLMs) have achieved remarkable success in vision-language tasks but their remote sensing (RS) counterpart are relatively under explored. Unlike natural images, RS imagery presents unique challenges that current MLLMs struggle to handle, particularly in visual grounding and spatial reasoning. This study investigates the limitations of CLIP-based MLLMs in RS, highlighting their failure to differentiate visually distinct yet semantically similar RS images. To address this, we introduce a remote sensing multimodal visual patterns (RSMMVP) benchmark. It is designed to evaluate MLLMs in RS tasks by identifying the CLIP-blind pairs, where CLIP-based models incorrectly assign high similarity scores to visually distinct RS images. Through a visual question answering (VQA) evaluation, we analyze the performance of state-of-the-art MLLMs, revealing significant limitations in RS specific representation learning. The results provide valuable insights into the weaknesses of CLIP-based visual encoding and offer a foundation for future research to develop more effective MLLMs tailored for remote sensing applications. Dataset is publicly available at \url{https://huggingface.co/datasets/IVUlab/RSMMVP}.
\end{abstract}

\section{Introduction}
\label{sec:Intro}
\begin{figure}[t]
      \centering
      \includegraphics  [width=0.5\textwidth]{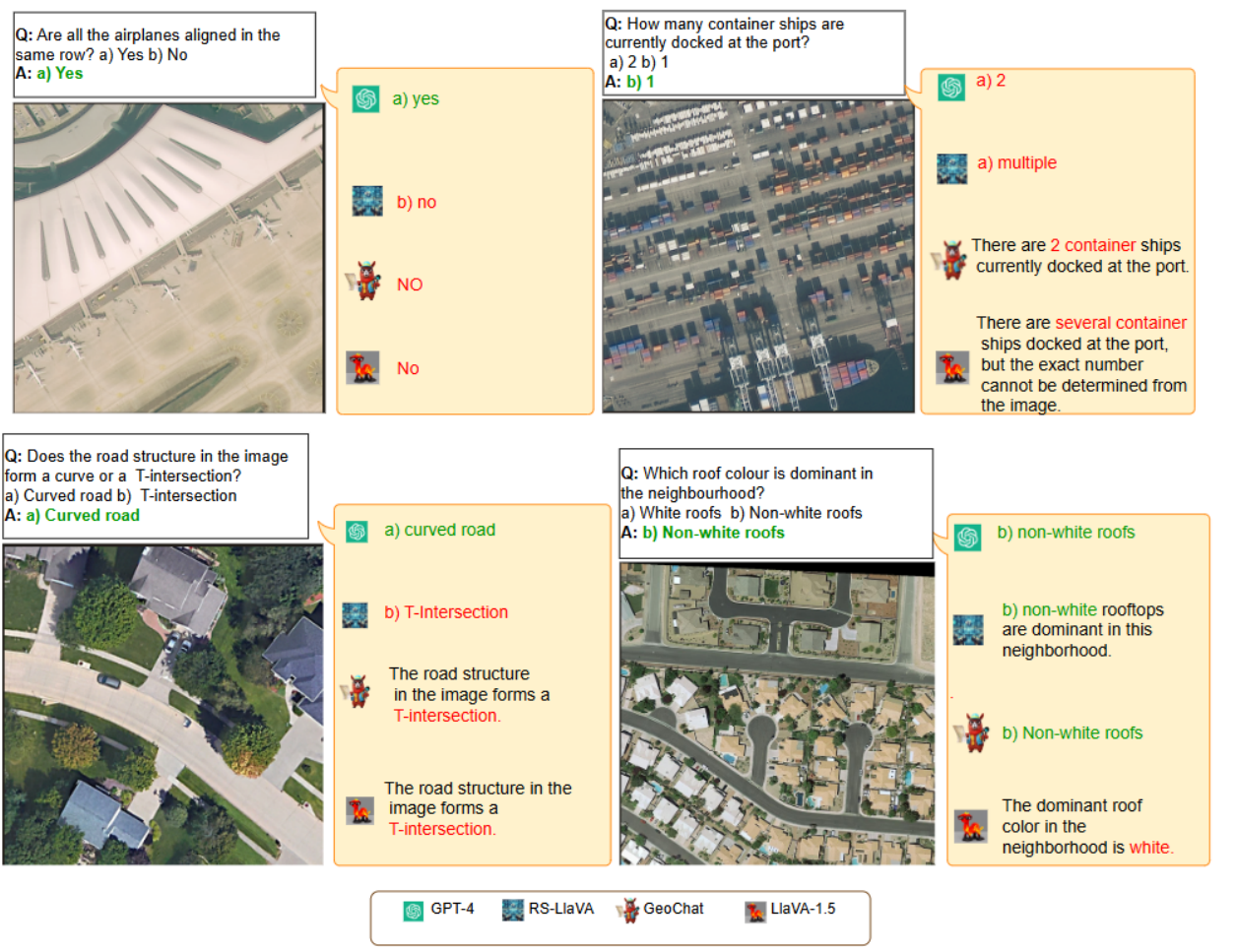}
      \caption{Examples from our RSMMVP benchmark demonstrating MLLMs shortcomings in RS VQA tasks due to weaknesses in visual grounding and reasoning. Incorrect and correct answers are highlighted in \textcolor{red}{red} and \textcolor{green}{green} respectively. LLaVA 1.5 7B variant is used in these results.}
      \label{Figure1}
\end{figure}
\blfootnote{\noindent This is the short version of the work submitted in fulfillment of the requirements for the MSc program at AMMI/AIMS Senegal.}
The development of multimodal large language models (MLLMs) has led to substantial advancements in vision-language tasks, particularly in the domain of natural images. These models integrate computer vision and natural language processing (NLP), enabling applications such as image captioning, visual question answering (VQA), and cross-modal reasoning \cite{liu2023visual, liu2024improved}. However, despite their success in natural image analysis, their effectiveness in remote sensing (RS) imagery is relatively in its infancy.

Unlike natural images, RS data presents distinct challenges, including fine-grained spatial structures, handling multiple sensors/modalities, and environmental variations \cite{xiao2024foundation, 7902107, Sumbul_2018}. The complexity of RS imagery requires RS specific vision encoders, yet most MLLMs rely on vision encoders pre-trained on natural image datasets, such as the CLIP-based architectures \cite{radford2021learningtransferablevisualmodels}. Consequently, these models often struggle with visual grounding, object counting, and spatial reasoning in RS applications.

This raises a fundamental question: ``Are these limitations due to deficiencies in language modeling, or are they due to weaknesses in the visual representation learning within MLLMs?" Existing research indicates that MLLMs primarily inherit their shortcomings from vision encoders, that fail to capture fine-grained visual information \cite{tong2024eyes,tong2025cambrian}. This limitation is particularly problematic for RS tasks, where distinguishing small-scale objects, and geometric structures is critical, see Figure~\ref{Figure1}.

Although several domain-specific MLLMs, such as GeoChat \cite{kuckreja2024geochat} and RS-LLaVA \cite{Bazi2024RSLLaVAAL}, have attempted to address these challenges by incorporating RS specific training datasets, they still inherit the limitations of CLIP-based feature extraction. These limitations are exacerbated by the high level of detail and fine-grained region-specific information in RS images, which necessitate operating at high resolution and can result in limited generalization ability to different modalities. 

This paper investigates the visual representation limitations of CLIP-based MLLMs in RS applications. Inspired by \cite{tong2024eyes}, we revisit the concept of CLIP-blind pairs, where visually distinct RS images receive high similarity scores from CLIP-based models and low similarity scores from vision-only models such as DINOv2 \cite{oquab2023dinov2}. These discrepancies reveal systematic errors in CLIP-based feature extraction, which hinder the ability of MLLMs to accurately interpret RS imagery.
To evaluate such shortcomings in RS, we propose a RS multimodal visual patterns (RSMMVP) benchmark specifically designed to assess the MLLMs' ability to distinguish fine-grained differences in RS imagery. By constructing a VQA task based on these CLIP-blind pairs, we evaluate how well MLLMs can visually ground information in high-resolution RS images. Our benchmark serves as a foundation for future research, highlighting the limitations of current models and guiding the development of more robust MLLMs tailored for RS applications.
In summary, our contributions are as follows:
\begin{itemize}
    \item We introduce RSMMVP, a benchmark designed to evaluate the ability of MLLMs on a RS vision centric task.
    \item We conduct a systematic evaluation of state-of-the-art MLLMs, revealing critical limitations in their vision-language alignment and representation learning for RS.
    \item Our benchmark and analysis can drive progress in multimodal foundation models for geospatial applications.
\end{itemize}

\begin{figure*}[t]
    \centering    \includegraphics[width=\textwidth]{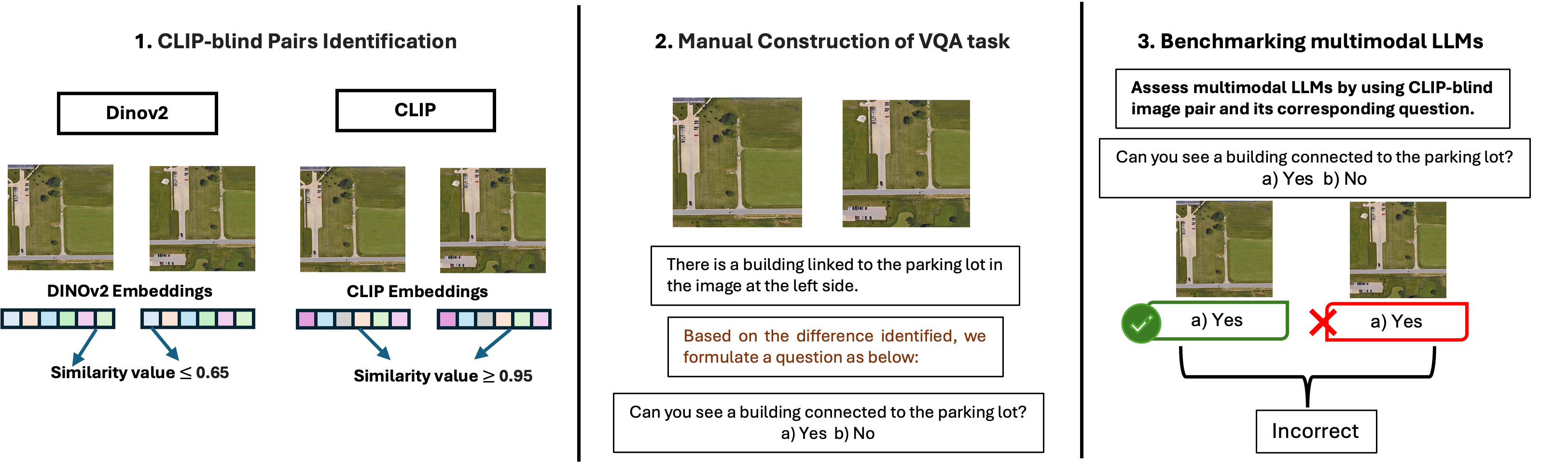}
    \caption{Constructing a benchmark for remote sensing imagery using CLIP-blind pairs. 
    From left to right: (1) CLIP-blind pairs identification
    (2) Manual construction of VQA task 
    (3) Query MLLMs with CLIP-blind pairs and evaluate responses. 
    A model is scored only if both answers for a CLIP-blind pair are correct.}
    \label{Figure2}
\end{figure*}

\section{Related Work}
\label{sec:Related}
We explore the limitations of MLLMs in processing RS imagery.
MLLMs leverage pre-trained LLMs \cite{liu2024llavanext, vicuna2023, jiang2023mistral7b} and integrate vision encoders \cite{radford2021learningtransferablevisualmodels, zhai2023sigmoidlosslanguageimage} to facilitate the alignment between visual and textual understanding. 
Language-visual alignment in such models can employ various techniques, such as simple projection layers\cite{liu2024improved}, Q-Formers \cite{li2023blip} and gated attention mechanisms \cite{alayrac2022flamingo}. Models like LLaVA and its variants \cite{liu2023visual, liu2024improved} highlight the role of high-quality instruction tuning data in enhancing multimodal learning for natural images. Unlike natural images, RS data exhibit fine-grained spatial structures in high resolution imagery, can include multiple modalities (e.g., hyperspectral data) and is impacted by environmental variations. Thus, the direct adaptation of MLLMs to RS data is challenging. Research on MLLMs for RS is largely based on LLaVA-based architectures, and several studies have used this framework to develop RS specific models, aiming to enhance vision-language alignment. GeoChat \cite{kuckreja2024geochat} and RS-LLaVA \cite{Bazi2024RSLLaVAAL} instruction-tune LLaVA models on satellite imagery datasets, leveraging domain-specific ones such as RSVQA \cite{lobry2020rsvqa}, FloodNet \cite{rahnemoonfar2021floodnet}, DOTA \cite{xia2018dota} and RSIVQA-DOTA \cite{zheng2021mutual}. Besides these studies, EarthGPT \cite{zhang2024earthgpt} incorporates Synthetic Aperture Radar (SAR) and infrared modalities, addressing the challenges of multi-sensor fusion in RS tasks. 
Although, GeoChat and RS-LlaVA models expand the scope of MLLMs for RS, yet their reliance on CLIP-based architectures limits their capacity to capture fine-grained spatial details and complex geospatial relationships.

Despite advancements in MLLMs, there remains a lack of systematic vision centric benchmarks specifically designed to evaluate vision-language capabilities in RS. Existing multimodal benchmarks, such as MMVP \cite{tong2024eyes}, POPE \cite{li2023evaluating}, and MM-Bench \cite{liu2024mmbench}, focus on evaluating robustness and multimodal reasoning but are primarily designed for natural image datasets rather than RS specific tasks. 
To address this gap, RSMMVP introduces a domain-specific benchmark that systematically evaluates MLLMs' ability to distinguish visually ambiguous CLIP-blind pairs in RS imagery. Unlike prior benchmarks, RSMMVP focuses on fine-grained visual reasoning in high-resolution RS data, providing a structured evaluation framework for assessing MLLM performance in such an application.
\section{Benchmark}
\label{sec:Method}
This section presents RSMMVP, a benchmark designed to evaluate the limitations of MLLMs in RS. The benchmark assesses how these models process RS imagery, particularly in cases where CLIP-based vision encoders fail to distinguish visually distinct image pairs. To address this, RSMMVP incorporates CLIP-blind pairs and construct a VQA dataset to analyze MLLM performance in RS specific tasks.

\noindent\textbf{CLIP-blind Pairs in Remote Sensing.} Inspired by previous work~\cite{tong2024eyes}, we define CLIP-blind pairs as image pairs that pertain high similarity scores from CLIP, while vision-only models assign them low similarity scores. As illustrated in Figure \ref{Figure2}, we first extract feature embeddings from the GeoChat training dataset \cite{kuckreja2024geochat}, selecting image pairs that exhibit inconsistencies between CLIP-based and vision-only representations. The embeddings are computed using CLIP-ViT-L/14 \cite{radford2021learningtransferablevisualmodels, dosovitskiy2020image} and DINOv2-ViT-L/14 \cite{oquab2023dinov2}. Cosine similarity scores are calculated for each image pair to quantify the differences between the two models' feature representations. An image pair is classified as CLIP-blind, if it receives a CLIP similarity score greater than 0.95, while having a DINOv2 similarity score below 0.6. These CLIP-blind pairs expose shortcomings in CLIP-based MLLMs.


\noindent\textbf{A Visual Question Answering Benchmark for Remote Sensing MLLMs.}
To evaluate MLLMs' ability to process RS imagery, we introduce a VQA benchmark specifically tailored for the identified CLIP-blind pairs. This benchmark 
serves as a proxy evaluation of MLLMs’ visual grounding abilities in high-resolution and complex RS imagery. The RSMMVP VQA dataset is constructed by analyzing 95 CLIP-blind image pairs, ensuring that it captures a diverse range of ambiguities and manually constructing the respective questions and choices. The benchmark consists of 300 questions within the VQA task, each designed to evaluate the models' ability to distinguish fine-grained differences in RS images. 

In order to ensure that the questions are not ambiguous we go through an iterative refinement process, where we use different human participants to answer the questions and evaluate their performance. This is followed by improving the questions' design that have proven the most difficult to multiple participants. This is conducted several times until mistakes are not consistent among participants and is mostly related to the difficulty level of the RS task. The human evaluation converged to 91.7\% accuracy from six participants, reflecting the challenge in such high resolution imagery with high level of detail.
The benchmark serves as a foundation for future research in vision-language adaptation for RS data, and encourages the development of domain-specific MLLMs that are better equipped to process geospatial high resolution imagery. 


\section{Experiments}
\label{sec:Exp}
This section presents our evaluation methodology, including benchmarking protocol, and MLLMs' performance analysis on RSMMVP, with focus on visual grounding, spatial reasoning, and object counting in RS imagery.

\noindent\textbf{Evaluation Protocols and Metrics.}
We adopt a paired question-answer assessment framework to ensure a rigorous evaluation of model accuracy. A model’s prediction was considered correct only if it accurately answered both questions associated with an image pair. 
This evaluation standard provides a robust measure of the ability of MLLMs  to distinguish fine-grained visual differences in RS imagery. We use the 300 image-question pairs in our RSMMVP dataset and evaluate the accuracy of each model using GPT-4o.


\begin{figure*}[t]
    \includegraphics[width=0.98\textwidth]{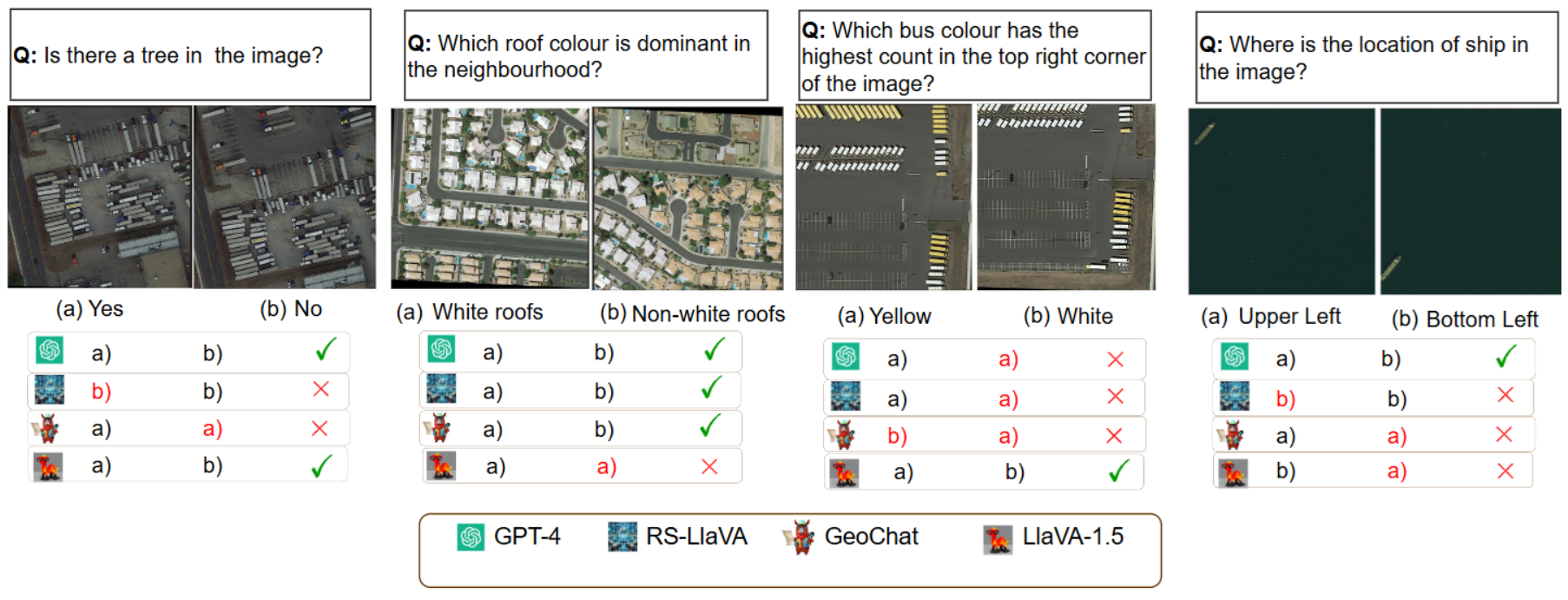}
\caption{Examples of questions from the RSMMVP benchmark. Incorrect answers are in \textcolor{red}{red}. A model is considered correct only if it answers both questions in a pair correctly. LLaVA 1.5 7B variant is used in these results. }
\label{Figure4}
\end{figure*}

\begin{figure}[t]
      \centering
      \includegraphics [width=0.5\textwidth]{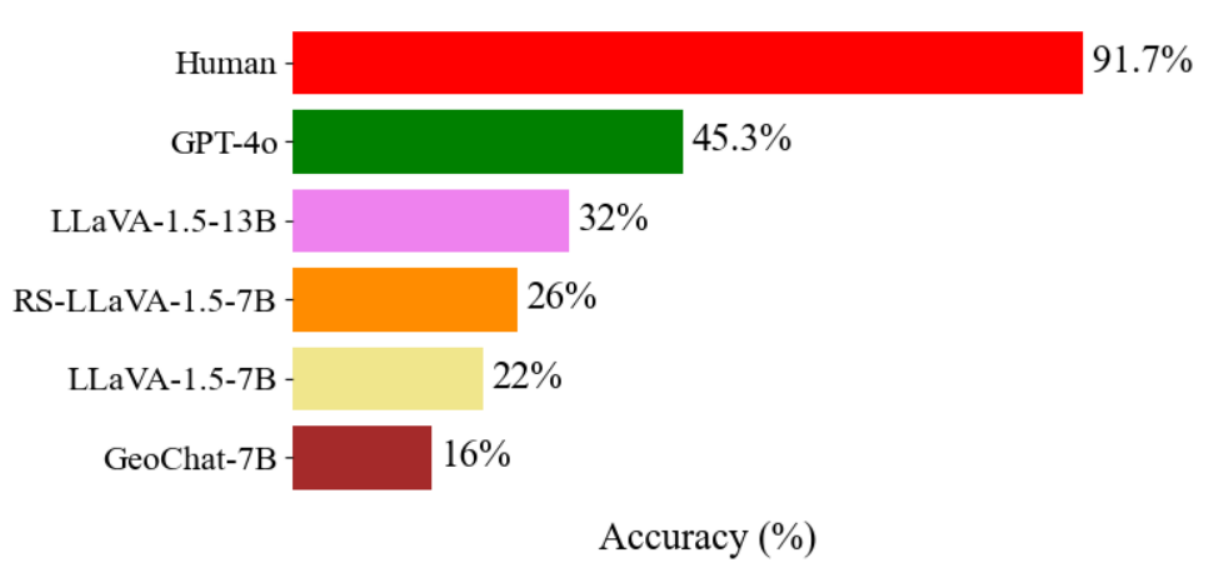}
      \caption{RSMMVP benchmark results of the  current  MLLMs with respect to the human performance.}
      \label{Figure3}
      \vspace{-2em}
\end{figure}

\noindent\textbf{Benchmarking MLLMs on RSMMVP.}
To thoroughly evaluate the ability of MLLMs in processing RS images, we conduct a comprehensive benchmarking study on the RSMMVP dataset. The evaluation results in Figure~\ref{Figure3} demonstrate a significant gap between human performance and the MLLMs' performance. 
Human participants achieved an accuracy of 91.7\%, which serves as an upper bound for the performance on our RSMMVP dataset. 
Among the models, GPT-4o (45.3\%) achieves the highest accuracy, yet it shows significant performance gap with respect to the human performance. 
In contrast,
the domain-adapted model RS-LLaVA-1.5-7B (26\%) exhibits slight improvements over the general-purpose counter-part LLaVA-1.5-7B (22\%), but still falls short in delivering robust performance for RS-specific applications.
Notably, GeoChat-7B (16\%) showed the lowest accuracy among the domain-adapted models. 
Note that the reported result for GeoChat corresponds to a re-trained version of the model, excluding the images used in constructing the benchmark.
This result suggests that despite being fine-tuned on RS datasets, GeoChat lacks generalization capabilities.
Finally, the LLaVA-1.5-13B achieved the second best result at 32\%.


Qualitative examples are shown in Figures~\ref{Figure1} and \ref{Figure4}
highlighting systematic errors made by these MLLMs.
These models consistently fail to capture the fine-grained visual details required for accurate RS analysis because of their reliance on CLIP vision encoder.
Although GeoChat and RS-LLaVA models are specifically tailored for RS applications, they still inherit the same underlying CLIP-based vision architectures limitations.

\begin{figure}[t]
      \centering
      \includegraphics [width=0.48\textwidth]{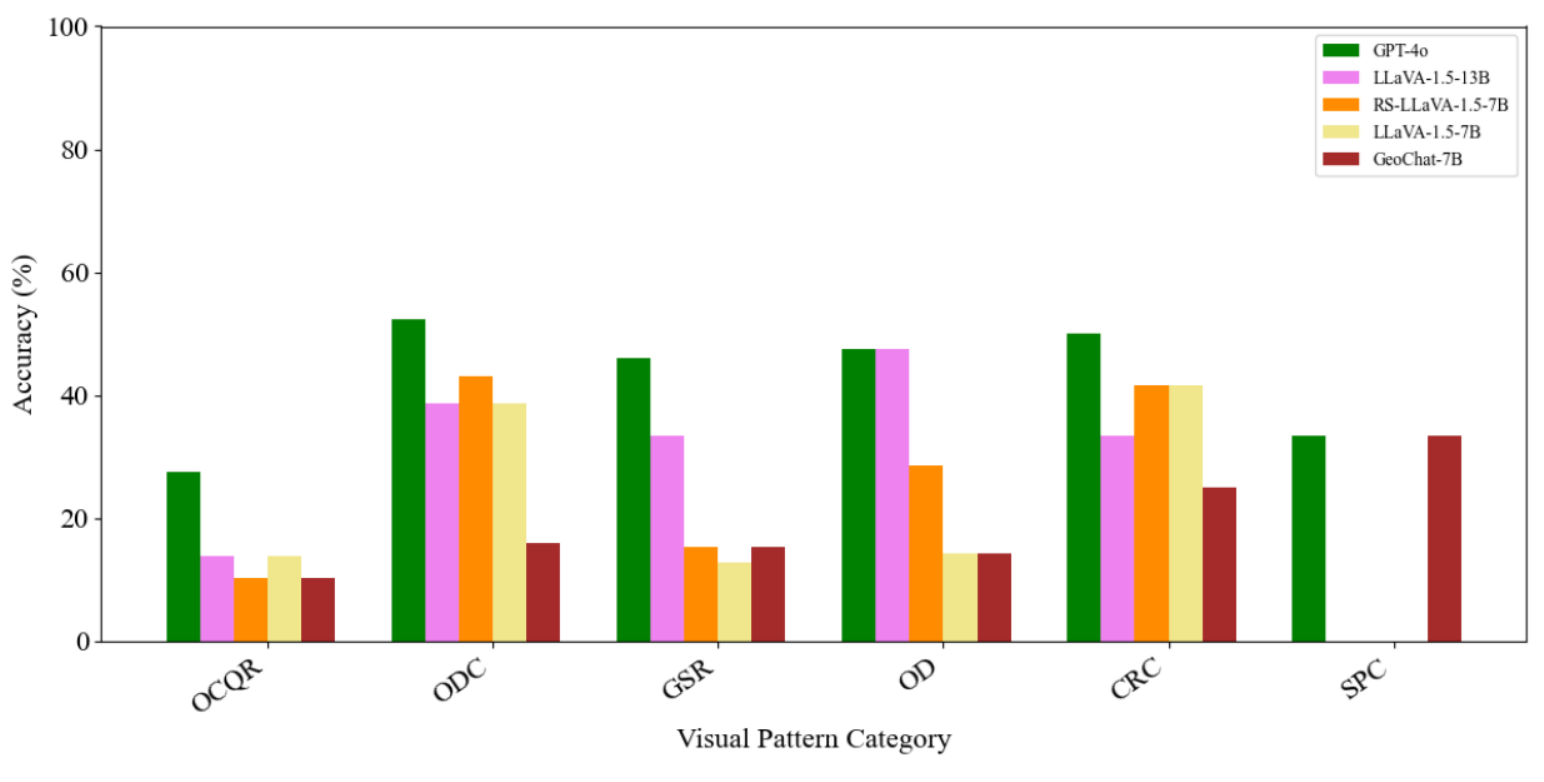}
      \caption{Fine-grained performance of current CLIP-based MLLMs on visual patterns in the RSMMVP benchmark. The evaluated visual pattern categories include Object Counting and Quantity Recognition (OCQR), Object Detection and Classification (ODC), Geometric and Spatial Relationships (GSR), Orientation and Directionality (OD), Color Recognition and Classification (CRC), and Size and Proximity Comparisons (SPC).}
      \label{Figure5}
      \vspace{-1.2em}
\end{figure}

\noindent\textbf{Visual Patterns in RS-CLIP-blind Pairs.}
Beyond quantitative evaluation, we conduct an error analysis to identify failure patterns in MLLMs when interpreting RS CLIP-blind pairs. To categorize these errors, we use question-answer analysis with GPT-4o \cite{openai2024gpt4technicalreport} and classify the most common visual patterns that lead to misinterpretations in CLIP-based models. This categorization offers deeper insights into the inherent limitations of these models and highlights the fundamental challenges in adapting CLIP-based architectures for RS applications.
Through this analysis, we identify six visual patterns in CLIP-based MLLMs when processing RS imagery which are provided with their respective proportion in the dataset: 
(i) Object Counting and Quantity Recognition (19.3\%).
(ii) Object Detection and Classification (29.3\%). 
(iii) Geometric and Spatial Relationships (26\%).
(iv) Orientation and Directionality (14\%). 
(v) Color Recognition and Classification (8\%). 
(vi) Size and Proximity Comparisons (2\%). 
In addition to four questions that were not categorized under any of the aforementioned visual patterns and were labelled as `Others'. In Figure~\ref{Figure5}, we illustrate the performance of various MLLMs on these visual patterns, offering a fine-grained assessment of their visual reasoning capabilities. GPT-4o outperforms other models in the visual patterns consistently. Overall, the models perform relatively better in color recognition and classification tasks, followed by object detection and classification.

\section{Conclusion}
\label{sec:Con}
This study introduced RSMMVP, a benchmark designed to evaluate the limitations of MLLMs in RS. By focusing on CLIP-blind pairs and designing a VQA task, we assessed the ability of MLLMs to distinguish challenging RS imagery.  
Our results reveal significant performance gaps, particularly in 
visual grounding and spatial reasoning, with significant weaknesses in object counting and quantity recognition as well as geometric and spatial relationships. 
 These findings demonstrate that current CLIP-based vision encoders struggle to capture RS specific features. 
These results emphasize the need for models specifically adapted to RS, capable of handling the unique complexities of RS data. 
Our benchmark serves as a foundation for future research to guide the development of more robust, RS specific vision language models.

\section*{Acknowledgments}
We acknowledge the support of the Natural Sciences and Engineering Research Council of Canada (NSERC) and Google research credits.

{\small
\bibliographystyle{ieeenat_fullname}
\bibliography{main}
}

\end{document}